\pdfoutput=1

\documentclass{article}

\usepackage{PRIMEarxiv}

\usepackage[utf8]{inputenc} 
\usepackage[T1]{fontenc}    
\usepackage{hyperref}       
\usepackage{url}            
\usepackage{booktabs}       
\usepackage{amsfonts}       
\usepackage{nicefrac}       
\usepackage{microtype}      
\usepackage{lipsum}
\usepackage{fancyhdr}       
\usepackage{graphicx}       
\graphicspath{{media/}}     

\usepackage{graphicx}
\usepackage{pgf}

\usepackage{tikz}
\usetikzlibrary{matrix,positioning}
\usetikzlibrary{shapes,snakes}

\tikzset{
  tab/.style={inner sep=0pt,
    nodes={inner sep=.333em,
      minimum height={\baselineskip+0.666em}
    }
  },
  vtab/.style={matrix of nodes,tab,
    row sep=-\pgflinewidth,column sep=-\pgflinewidth,
    nodes in empty cells,
    nodes={draw,align=left,text width=#1}
  },
  vtab/.default=3cm,
  htab/.style={matrix of nodes,draw,tab,anchor=north west},
  every edge/.append style={font=\footnotesize\strut,inner ysep=.1em},
  pfeil/.style={out=270,in=90,->}
}

\usepackage{caption}
\usepackage{subcaption}

\usepackage{multirow}

\usepackage[ruled,vlined]{algorithm2e}
\usepackage{amsmath}
\usepackage{todonotes}
\usepackage{amssymb}

\usepackage{oubraces}

\newtheorem{definition}{Definition}

\title{Neural RELAGGS
}

\author{
  Lukas Pensel, Stefan Kramer \\
  Institut f\"ur Informatik \\
  Johannes Gutenberg-Universit\"at \\
  Mainz\\
  \texttt{pensel@uni-mainz.de,kramer@informatik.uni-mainz.de} \\
}

\begin{document}
\maketitle

\begin{abstract}
Multi-relational databases are the basis of most consolidated data collections in science and industry today. Most learning and mining algorithms, however, require data to be represented in a propositional form. While there is a variety of specialized machine learning algorithms that can operate directly on multi-relational data sets, propositionalization algorithms transform multi-relational databases into propositional data sets, thereby allowing the application of traditional machine learning and data mining algorithms without their modification. One prominent propositionalization algorithm is RELAGGS by Krogel and Wrobel, which transforms the data by nested aggregations. We propose a new neural network based algorithm in the spirit of RELAGGS that employs
trainable composite aggregate functions instead of the static aggregate functions used in the original approach. In this way, we can jointly train the propositionalization with the prediction model, or, alternatively, use the learned aggegrations as embeddings in other algorithms. We demonstrate the increased predictive performance by comparing N-RELAGGS with RELAGGS and multiple other state-of-the-art algorithms.
\end{abstract}

\keywords{Network architecture \and Propositionalization \and Relational data mining \and Aggregation}

\section{Introduction}
\label{sec:intro}

While neural networks are among the most used and successful machine learning algorithms at the time, they mostly rely on propositional, i.e. single table, input data. Unfortunately, many real-world problems are based on relational data structures, such as social networks, customer accounts, and health records. There are multiple approaches to incorporate those relational structures into the topology of the neural network, such as relational neural networks \cite{LRNN}, and there are even more algorithms to transform a relational data set into a propositional data set, such as RELAGGS \cite{propositionalization:relaggs}, in order to make it usable for all traditional machine learning and data mining algorithms.

The goal of this work is to merge these approaches and build a propositionalization algorithm into the topology of a neural network. We aim to combine the simplicity of using propositional data as input for our predictions with the possibility to adjust the way in which the data is transformed, in order to optimize the propositionalization for the specific prediction task while also training the predictor.

Along those lines, we introduce an approach to trainable aggregate functions based on the composition of parameterized mappings and unparameterized aggregate functions. We show further that we can use those functions to jointly propositionalize and predict relational data in a neural network context and we provide an implementation of such an approach with our Neural RELAGGS (N-RELAGGS) algorithm. Lastly, we evaluated our algorithm and compared it with other state-of-the-art propositionalization methods on multiple benchmark data sets of varying size. 

The main contributions of this work are:
\begin{itemize}
    \item We introduce aggregate functions for propositionalization, in the style of RELAGGS, that are {\em trainable} and can thus be further optimized in learning. The trainable aggregate functions are composite functions and consist of aggregation layers in neural networks.
    \item We implement those trainable aggregate functions and provide them in a joint propositionalization and prediction algorithm called N-RELAGGS (Neural RELAGGS). N-RELAGGS is provided with an interface to the python-rdm\footnote{https://github.com/xflows/rdm/} library and can thus be compared and combined with other algorithms there.
    \item Experimental results show that N-RELAGGS compares favorably to the compared algorithms, in particular on larger data sets, and that it has a significantly higher prediction performance than most compared algorithms.
    \item We present a larger dataset derived from the DBLP-Citation-network as a novel relational database benchmark for propositionalization and relational learning methods. Experiments show that N-RELAGGS outperforms RELAGGS on this database as well.
\end{itemize}

This paper is organized as follows: In Section \ref{sec:related_work} we give a short introduction to relational data mining, introduce the approach of propositionalization and present the algorithms used for comparison in this work. The idea of composite aggregate functions and our implementation of them using neural networks is described in Section \ref{sec:methods}. The experimental setup and the data sets we use are specified in Section \ref{sec:experiments} together with the achieved performance. In Section \ref{sec:discussion} we discuss our results and we present our conclusions and an outlook on further work in Section \ref{sec:conclusion}.
\section{Related work}
\label{sec:related_work}
In this section we will give a short overview of the general problem of relational data mining and introduce the propositionalization approach to relational data mining. Additionally. we introduce the algorithms used for comparison in our experiments.

\subsection{Relational Data Mining}

\paragraph{} 
Relational data mining describes the task of finding patterns among multiple connected data tables, instead of just one single table, as with traditional data mining. Therefore, traditional algorithms cannot be directly applied to this task, which creates the necessity of specialized algorithms and methods to handle relational data. One class of such algorithms are propositionalization algorithms, which are the focus of this work.

\subsection{Propositionalization}

\paragraph{} 
Propositionalization is the  representation change of  transforming a  relational representation of a learning problem into a propositional (feature-based, attribute-value) representation \cite{propositionalization:kramer}. The idea is to build higher-level feature representations from lower-level relational data, just like super-pixels are constructed from pixels in image data. Taking such an approach, feature construction can be decoupled from model construction. In propositionalization, the search space of relational features is not gradually searched and expanded as in inductive logic programming (ILP), but all relational features with certain properties, e.g., up to some maximal syntactic size or within a minimal and maximal frequency of occurrence, are generated and used to transform the representation. The advantage is that it is possible to take advantage of any progress with propositional learning algorithms in this way. The disadvantage is the potential loss of information due to size or frequency constraints. If propositionalization does not give the desired results, it should at least be used as a baseline to show that more advanced search and optimization strategies are worth the effort.

\begin{table}[!htbp]
    \centering
    \begin{tabular}{rclr}
        ? & - & person(K), parent(K, Y), has\_pet(Y, cat). & (a) \\
        \cline{1-4}
         p(K, Z) & :- & person(K), has\_account(K, Y), overdraft(Y, Z). & (b) 
    \end{tabular}
    \caption{Propositionalization: (a) A conjunctive query with key K defining a Boolean feature for an instance K (true if it succeeds for some K, false otherwise). This is called an existential feature. (b)  A clause defining a feature for a relational example. K denotes the key (i.e., the identifier of the instance). It is assumed that a person may have several accounts Y, each with a different numerical overdraft limit Z. Such a clause may be the basis for several types of features, e.g., by applying aggregate functions to the set of different values for Z. This is called an aggregate feature.}
    \label{tab:propositionalization}
\end{table}

\paragraph{}
Propositionalization schemes can be categorized according to the types of features that are constructed. One basic distinction is the one between existential features and aggregate features (see Table \ref{tab:propositionalization}). Existential features are defined by conjunctive queries, which, when succeeding for an instance, give the value true, and false, otherwise. Variants are possible by counting the number of successful proofs.
Aggregate features \cite{propositionalization:relaggs} are more complex: We consider a defined set of variables except the key, which give the answer substitutions for that query (when the key variable is bound to some instance identifier). In the above example (see Table \ref{tab:propositionalization} (b)), K is the key, and the user has defined variable Z to be the one of interest for relational feature construction. When the query is evaluated for some instance K, we gather all values for Z , and, in the final step, apply a user-defined set of aggregate functions (like minimum, maximum, mean, standard deviation, mode, etc.) to that set of values to define propositional values. Clearly, variants are possible: The clause in Table \ref{tab:propositionalization} (b) can be the basis for some test against a threshold (e.g., $Z > 10000$), or it can be used not to turn the problem into a propositional problem, but, without aggregates, into a multi-instance learning or multi-tuple learning problem \cite{de2008logical}.

\subsubsection{RELAGGS}

\paragraph{} 
The RELAGGS, or relational aggregation, algorithm \cite{propositionalization:relaggs} is built on aggregate features as described above and transforms a relational data representation into a propositional one by nested aggregation. Given an entry of a relational data table $v$ and corresponding entries $W$ of a related table, RELAGGS transforms $v$ and $W$ into $v \oplus max(W) \oplus \dots \oplus std(W)$, where $\oplus$ represents the concatenation of two vectors. By iteratively aggregating connected entries of related tables, the algorithm converts a relational database into a propositional data table. Commonly used aggregation functions for this algorithm are average, maximum, minimum, standard deviation, and sum. Since this propositionalization approach is quite simple, we chose to build our approach based upon its structure.

\subsubsection{Other comparison algorithms}

\paragraph{Aleph} One of the most popular ILP systems is Aleph \cite{srinivasan2001aleph}. Given a background knowledge, positive examples and negative examples, it generates a set of clauses, which can be used as features, and therefore yields a propositional representation of the data.

\paragraph{RSD} In order to discover relational subgroups, the RSD algorithm \cite{rsd} generates a propositional representation of the provided relational database in a first step. This representation can be directly used with other algorithms.

\paragraph{TreeLiker} The ReIF algorithm \cite{treeliker}, part of the TreeLiker software, generates a propositional data representation of a relational database, by constructing a set of tree-like features.

\paragraph{Wordification}
Wordification \cite{wordification_1} is an approach to propositionalization based on text mining methods. It transforms a relational database into a representation akin to a Bag-Of-Words representation. 
\section{Methods}
\label{sec:methods}
Here we introduce the idea of composite aggregate functions as trainable aggregate functions and present a possible implementation of those in the form of Neural RELAGGS (N-RELAGGS).

\begin{table}[!htbp]
    \centering
    \begin{tabular}{c|c}
         Name & description \\
         \hline
         $n$ & the number of data instances in a batch or data set \\
         $l$ & the number of features \\
         $C$ & a set of collections of data \\
         $m$ & the number of items in a collection of data \\
         $k$ & the number of base aggregate functions \\
         $\Lambda$ & a dense neural network layer \\
    \end{tabular}
    \caption{Nomenclature of used variables and parameters.}
    \label{tab:nomenclature}
\end{table}

\subsection{Learned aggregation}

Preset aggregate functions are strongly limited in the amount of information they can retain. Additionally, those functions do not have trainable parameters to fit the aggregate to a specific task. In order to improve the amount of important information for a specific task, we want to construct trainable aggregate functions. 

\begin{definition}[Aggregate function]
An aggregate function $\psi$ maps a collection of elements of an input domain $I$ to a single element of an output domain $O$.
\begin{align*}
    \psi \colon \bigcup_{n \in \mathbb{N}} I^n &\to O
\end{align*}
\end{definition}

\begin{definition}[Trainable aggregate function]
A trainable aggregate function $\psi_\theta$ is an aggregate function for each possible set of parameters $\theta \in \Theta$.
\begin{align*}
    \psi_\theta \colon \bigcup_{n \in \mathbb{N}} I^n \to O
\end{align*}
\end{definition}

\begin{definition}[Composite aggregate function]
Given a base aggregate function $\psi$ for the input domain $I^*$ and the output domain $O^*$ and two functions $f_{\theta_1}$ and $g_{\theta_2}$ with
\begin{align*}
    f_{\theta_1} &\colon I \to I^*\\
    g_{\theta_2} &\colon O^* \to O
\end{align*}
we obtain the composite aggregate function $\psi_{(\theta_1,\theta_2)} = g_{\theta_2} \circ \psi \circ f_{\theta_1}$, which is a trainable aggregate function.
\begin{align*}
  \psi_{(\theta_1,\theta_2)} \colon \bigcup_{n \in \mathbb{N}} I^n &\to O \\
       s &\mapsto g_{\theta_2}(\psi(f_{\theta_1}(s)))
\end{align*}
\end{definition}

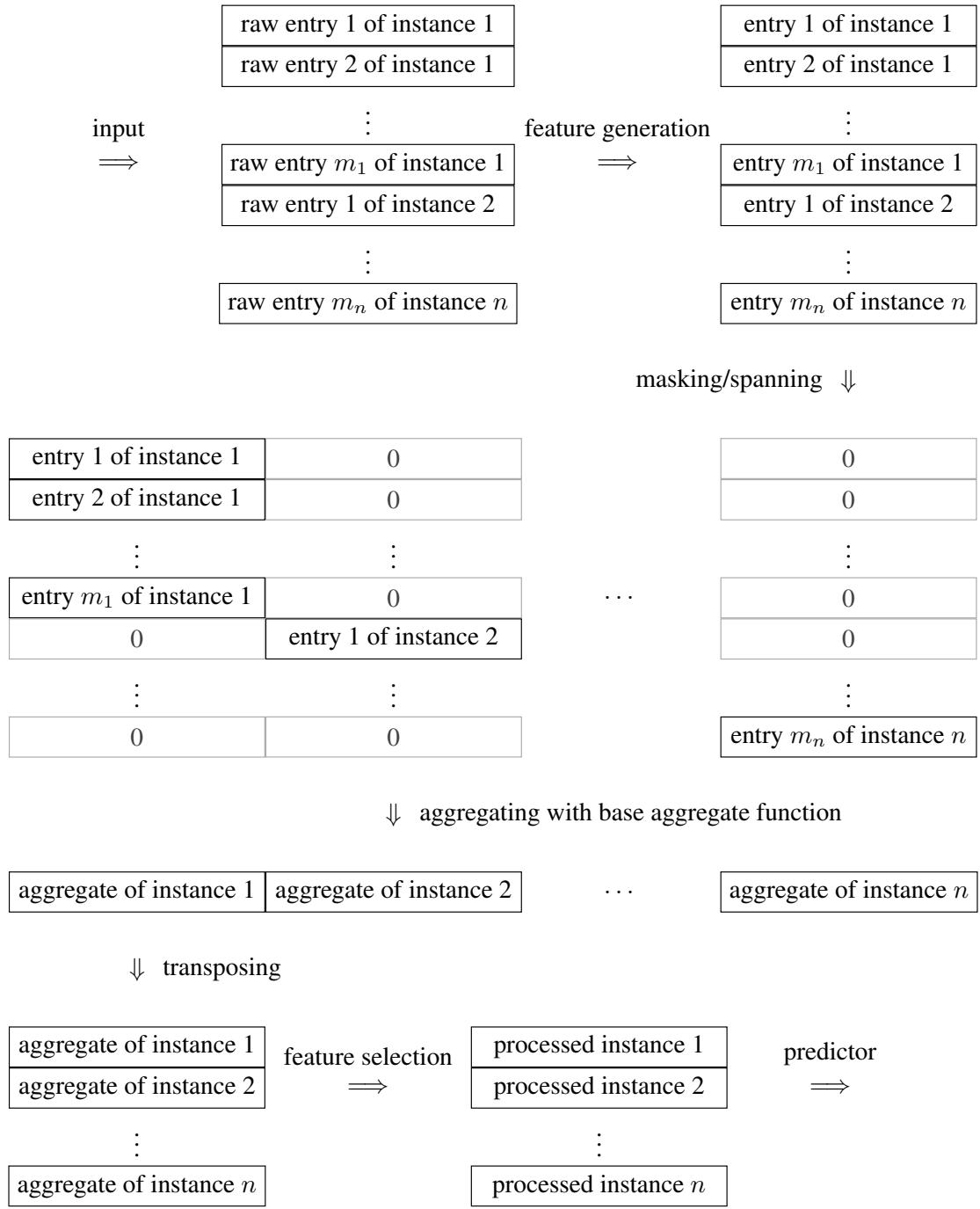
\begin{figure}[!htbp]
    \centering
    \resizebox{0.9\textwidth}{!}{
    \begin{tikzpicture}[
    textnode/.style={rectangle,draw=black,minimum width=3.5cm,minimum height=0.5cm}, 
    textnode2/.style={rectangle,draw=black,minimum width=4cm,minimum height=0.5cm}, 
    dotsnode/.style={rectangle},
    fadenode/.style={rectangle,draw=black!33,minimum width=3.5cm,minimum height=0.55cm}
    ]
    \node[textnode2] (in11) {raw entry 1 of instance 1};
    \node[textnode2] (in12) [below= 0cm of in11] {raw entry 2 of instance 1};
    \node[dotsnode] (in1d) [below= 0cm of in12] {$\vdots$};
    \node[textnode2] (in1n) [below= 0cm of in1d] {raw entry $m_1$ of instance 1};
    \node[textnode2] (in21) [below= 0cm of in1n] {raw entry 1 of instance 2};
    \node[dotsnode] (ind) [below= 0cm of in21] {$\vdots$};
    \node[textnode2] (innn) [below= 0cm of ind] {raw entry $m_n$ of instance $n$};
    
    \node[dotsnode] (in2pre) [right= of in1n] {$\Longrightarrow$};
    \node[dotsnode] (in2pre2) [above= 0cm of in2pre] {feature generation};
    
    \node[dotsnode] (input) [left= of in1n] {$\Longrightarrow$};
    \node[dotsnode] (input2) [above= 0cm of input] {input};
    
    \node[textnode] (pre1n) [right= of in2pre] {entry $m_1$ of instance 1};
    \node[dotsnode] (pre1d) [above= 0cm of pre1n] {$\vdots$};
    \node[textnode] (pre12) [above= 0cm of pre1d] {entry 2 of instance 1};
    \node[textnode] (pre11) [above= 0cm of pre12] {entry 1 of instance 1};
    \node[textnode] (pre21) [below= 0cm of pre1n] {entry 1 of instance 2};
    \node[dotsnode] (pred) [below= 0cm of pre21] {$\vdots$};
    \node[textnode] (prenn) [below= 0cm of pred] {entry $m_n$ of instance $n$};
    
    \node[dotsnode] (pre2mask) [below=0.5cm of prenn] {$\Downarrow$};
    \node[dotsnode] (pre2mask2) [left= 0cm of pre2mask] {masking/spanning};

    \node[fadenode] (mask311) [below=0.5cm of pre2mask] {\textcolor{black!75}{0}};
    \node[fadenode] (mask312) [below= 0cm of mask311] {\textcolor{black!75}{0}};
    \node[dotsnode] (mask31d) [below= 0cm of mask312] {$\vdots$};
    \node[fadenode] (mask31n) [below= 0cm of mask31d] {\textcolor{black!75}{0}};
    \node[fadenode] (mask321) [below= 0cm of mask31n] {\textcolor{black!75}{0}};
    \node[dotsnode] (mask3d) [below= 0cm of mask321] {$\vdots$};
    \node[textnode] (mask3n) [below= 0cm of mask3d] {entry $m_n$ of instance $n$};
    
    \node[dotsnode] (mask2mask) [left= of mask31n] {$\dots$};
    
    \node[fadenode] (mask21n) [left=  of mask2mask] {\textcolor{black!75}{0}};
    \node[dotsnode] (mask21d) [above= 0cm of mask21n] {$\vdots$};
    \node[fadenode] (mask212) [above= 0cm of mask21d] {\textcolor{black!75}{0}};
    \node[fadenode] (mask211) [above= 0cm of mask212] {\textcolor{black!75}{0}};
    \node[textnode] (mask221) [below= 0cm of mask21n] {entry 1 of instance 2};
    \node[dotsnode] (mask2d) [below= 0cm of mask221] {$\vdots$};
    \node[fadenode] (mask2n) [below= 0cm of mask2d] {\textcolor{black!75}{0}};
    
    \node[textnode] (mask11n) [left= 0cm of mask21n] {entry $m_1$ of instance 1};
    \node[dotsnode] (mask11d) [above= 0cm of mask11n] {$\vdots$};
    \node[textnode] (mask112) [above= 0cm of mask11d] {entry 2 of instance 1};
    \node[textnode] (mask111) [above= 0cm of mask112] {entry 1 of instance 1};
    \node[fadenode] (mask121) [below= 0cm of mask11n] {\textcolor{black!75}{0}};
    \node[dotsnode] (mask1d) [below= 0cm of mask121] {$\vdots$};
    \node[fadenode] (mask1n) [below= 0cm of mask1d] {\textcolor{black!75}{0}};

    \node[dotsnode] (mask2agg) [below=0.5cm of mask2n] {$\Downarrow$};
    \node[dotsnode] (mask2agg2) [right=0cm of mask2agg] {aggregating with base aggregate function};
    
    \node[textnode] (agg2) [below=0.5cm of mask2agg] {aggregate of instance 2};
    \node[textnode] (agg1) [left= 0cm of agg2] {aggregate of instance 1};
    \node[dotsnode] (aggd) [right= of agg2] {$\dots$};
    \node[textnode] (aggn) [right= of aggd] {aggregate of instance $n$};
    
    \node[dotsnode] (agg2out) [below=0.5cm of agg1] {$\Downarrow$};
    \node[dotsnode] (agg2out2) [right= 0cm of agg2out] {transposing};

    \node[textnode] (out1) [below=0.5cm of agg2out] {aggregate of instance 1};
    \node[textnode] (out2) [below=0cm of out1] {aggregate of instance 2};
    \node[dotsnode] (outd) [below=0cm of out2] {$\vdots$};
    \node[textnode] (outn) [below=0cm of outd] {aggregate of instance $n$};
    
    \node[dotsnode] (out2post) [right= of out2] {$\Longrightarrow$};
    \node[dotsnode] (out2post2) [above=0cm of out2post] {feature selection};
    
    \node[textnode] (post2) [right= of out2post] {processed instance 2};
    \node[textnode] (post1) [above= 0cm of post2] {processed instance 1};
    \node[dotsnode] (postd) [below=0cm of post2] {$\vdots$};
    \node[textnode] (postn) [below=0cm of postd] {processed instance $n$};

    \node[dotsnode] (output) [right= of post2] {$\Longrightarrow$};
    \node[dotsnode] (output2) [above=0cm of output] {predictor};

    \end{tikzpicture}
    }
    
\caption{Schematic of the neural network based composite aggregate function. We take $n$ instances with $m_1, \dots, m_n$ entries as input and each entry is passed through a feature generation network layer. Then the tensors are spanned in order to aggregate the entries of singular instances. The resulting tensor of aggregates is passed through a feature selection network layer and subsequently forwarded to the predictor model.}
    \label{fig:nrelaggs}
\end{figure}

In this work, we integrate the aggregate functions into a neural network, building a composite aggregate function consisting of neural network layers and some base aggregate function. Therefore, we can jointly learn the aggregate function and train the model. The schematic of this approach is depicted in Figure \ref{fig:nrelaggs}. The method design is inspired by global one-dimensional pooling layers, where sequences of data are combined into a single representation. The nomenclature of variables and parameters is gathered in Table \ref{tab:nomenclature}. In order to use such a global pooling layer to aggregate a batch of $n$ collections $C$ of data, with collection $c_i \in C$ consisting of $m_i = |c_i|$ data entries with $l$ features, we need to transform the data into a three-dimensional tensor with the shape $(n,l,\Bar{m} = \sum_{i<n}m_i)$. This is achieved by concatenating all collections into a tensor of the shape $(\Bar{m},l,1)$ and multiplying it with a mask $M$ of the shape $(\Bar{m},1,n)$, resulting in a tensor of the shape $(\Bar{m},l,n)$. By transposing the tensor, i.e. reversing the order of the axis, we obtain the required shape. Applying global pooling to this returns a tensor of the shape $(n,l)$.

\begin{algorithm}
\SetAlgoLined
\caption{Composite neural network aggregation}
\label{algo:neural_agg}
\KwData{input data $X$ of shape $(\Bar{m},l,1)$, input mask $M$ of shape $(\Bar{m},1,n)$}
\KwResult{aggregated data $\Tilde{X}$ of shape $(n,\Bar{l})$}
$X^* = \Lambda_{\text{feature generation}}(X) $ \tcp*{$(\Bar{m},l,1) \to (\Bar{m},l^*,1)$}
$X^* = X^* \times M$ \tcp*{$(\Bar{m},l^*,1) \times (\Bar{m},1,n) \to (\Bar{m},l^*,n)$}
$\Tilde{X}^* =$ aggregate($X^*$) \tcp*{$(\Bar{m},l^*,n) \to (l^*,n)$}
$\Tilde{X}^* =$ transpose($\Tilde{X}^*$) \tcp*{$(l^*,n) \to (n,l^*)$}
$\Tilde{X} = \Lambda_{\text{feature selection}}(\Tilde{X}^*)$ \tcp*{$(n,l^*) \to (n,\Bar{l})$}
\end{algorithm}

In order to introduce trainable parameters to the aggregate function we add two dense network layers, as presented in Algorithm \ref{algo:neural_agg}. Before the aggregation we apply some sort of feature generation by transforming the input data with a dense network layer $\Lambda_{\text{feature generation}}$, where each input entry with $l$ features is mapped to a processed input entry with $l^*$ features. After the aggregation,  we add a feature selection layer $\Lambda_{\text{feature selection}}$, which maps the aggregates of shape $(n,l^*)$ to a selected output of shape $(n,\Bar{l})$.

In this manner, we can use one specific pooling method, such as average or max pooling. However, we can increase the expressiveness of the aggregate by incorporating multiple different pooling, i.e. aggregation, methods and concatenating the results, similar to the RELAGGS algorithm. So, if $k$ is the number of different pooling methods used for aggregation, we get the final sequence of tensors:

\begin{equation*}
    \overunderbraces{&&&&\br{3}{\text{aggregate}}}
    {&(\Bar{m},l) \rightarrow (\Bar{m},l^*)& \rightarrow &(\Bar{m},l^*,1) \times (\Bar{m},1,n) \rightarrow &(\Bar{m},l^*,n)& \rightarrow &(n,k \times l^*)& \rightarrow (n,\Bar{l})}%
    {&\br{1}{\text{feature generation}} &  &\br{2}{\text{span the tensors}} & &\br{2}{\text{feature selection}}}
\end{equation*}

\subsection{Neural RELAGGS}

Our goal is to transform a multi-relational data set, a relational database in star schema with provided target table and target attribute, by aggregating, using trainable aggregate functions, into a propositional representation and take this representation as input for a prediction network. 

In order to handle a multi-relational data set, we have to combine multiple aggregations into one network. The implementation of those aggregation layers is described in Section \ref{sec:agg_layer}. Additionally, we have to transform the multi-relational data set into a representation we can feed into a neural network and retrieve the structure of the relations, in order to build it into the network architecture. This transformation and the retrieval of the structure as aggregation plan $P$ is described in Section \ref{sec:data_process}.

The transformed data $X$ can now be aggregated according to the aggregation plan $P$, and the resulting propositional data representation is then fed into a prediction network $\Lambda_{pred}$, as described in this section, in Algorithm \ref{algo:nrelaggs} below. Since the whole network architecture is jointly implemented, it also can be jointly trained using regular optimizers, such as gradient descent or Adam \cite{network:adam}. 

To minimize the number of hyperparameters, we just set a feature generation factor and a feature selection factor to adjust the aggregation layers. They describe the factor by which the number of features is increased. A factor of 1.0 retains the number of features after the pass through the network layer, where a factor of 0.5 halves the number of features.

We implemented the N-RELAGGS algorithm in Python using TensorFlow \cite{tensorflow2015-whitepaper} as the underlying deep-learning framework and made the source code publicly available\footnote{https://github.com/kramerlab/n-relaggs}.

\begin{algorithm}
\SetAlgoLined
\caption{Neural RELAGGS}
\label{algo:nrelaggs}
\KwData{Relational databse in star schema $DB$, with target table $DB.target\_table$ and target attribute $DB.target\_table.target$}
\KwResult{Predicted values $y\_pred$}
$X = [X_{data},X_{ids}], P =$ database\_converter($DB$) \tcp*{Preprocess the database, see Algorithm \ref{algo:db2py}}
$\Tilde{X}_{data} = copy(X_{data})$\;
\ForAll{$p \in P$}
{
$X_{agg} = []$\;
\ForAll{$i \in p[0]$}{
$X_{agg}.append(\text{aggregation\_layer}_i(\Tilde{X}_{data}[i],X_{ids}[i]))$ \tcp*{Aggregate data, see Algorithm \ref{algo:agg_layer}}
}
\ForAll{$i \in p[1]$}{
$X_{agg}.append(X_{data}[i])$ \tcp*{Non-aggregated part of the combination}
}
$X_{agg} = concatenate(X_{agg})$\;
$\Tilde{X}_{data}[p[1]] = X_{agg}$ \tcp*{Store data for later combination}
}
$y\_pred = \Lambda_{pred}(\Tilde{X}_{data}[0])$ \tcp*{All data is aggregated into $\Tilde{X}_{data}[0]$}
\end{algorithm}

 \subsection{Aggregation layer}
 \label{sec:agg_layer}
 
Since there are only global average pooling and global max pooling layers implemented in TensorFlow, we would be limited to only use those two base aggregate functions. However, internally the pooling layers are implemented using dimension reducing operations, such as sum or mean. Therefore, we can employ more base aggregate functions by incorporating those operations directly instead of the pooling layers. As described above, we first transform the input into a tensor of the shape $(\Bar{m},l^*,n)$ by generating features and spanning it using mask $M$. Now the tensor is reduced along the first dimension, resulting in a tensor of shape $(l^*,n)$ and after transposing $(n,l^*)$. 

Unfortunately, this approach requires us to construct huge tensors in the spanning step, which cannot be stored in the systems memory for bigger data sets. Theoretically, the use of sparse matrices should eliminate this problem, but to our knowledge there exists no available implementation of genuine sparse tensor multiplication, where the result is a sparse tensor and internally everything is sparsely represented and not cast to a dense tensor.

To solve this problem, we incorporate the idea of using the TensorFlow segment operations as base aggregate functions \cite{larionov_2019}. Given a two-dimensional tensor $X$ of shape $(\Bar{m},l^*)$ and $\Bar{m}$ corresponding indices $J$, with $j <= n \forall j \in J$, relating each entry of $X$ to its data instance, the segment operation reduces $X$ along the first axis to a tensor of shape $(n,l^*)$, by combining all entries with the same index. Instead of computing an $(\Bar{m},l,n)$ tensor, we only need an $(\Bar{m},l^*)$ and an $(\Bar{m},1)$ tensor. This alters the sequence of tensors to
 
 \begin{equation*}
    \overunderbraces{&\br{2}{\text{feature generation}}&&\br{3}{\text{aggregate}}}%
    {&(\Bar{m},l) \rightarrow &(\Bar{m},l^*)& \rightarrow &(\Bar{m},l^*),(\Bar{m},1)& \rightarrow &(n,k \times l^*)& \rightarrow (n,\Bar{l}),}%
    { &&\br{3}{\text{combine data with indices}} & &\br{2}{\text{feature selection}}}
\end{equation*}
 
 and the implementation of this aggregation layer is described in Algorithm \ref{algo:agg_layer}.
 
 Since we incorporate the sum, maximum, minimum and average as base aggregates in our implementation, we can learn them by setting a neuron to pass through one unmodified value. Additionally, we can learn those base aggregates of combinations of features. For instance, the maximum, minimum, average or total difference between two features is calculated by assigning one feature the weight of 1, the other of $-1$ and all other features of 0. However, aggregate functions which require knowledge about the whole collection of data during the feature generation step, such as the standard deviation, cannot be exactly recreated.

\begin{algorithm}
\SetAlgoLined
\caption{Aggregation layer}
\label{algo:agg_layer}
\KwData{Data tensor $X$, indices tensor $J$}
\KwResult{Aggregated tensor $\Tilde{X}$}
$X^* = \Lambda_{in}(X) $ \tcp*{Feature generation layer}
$\Tilde{X}^* = []$\;
$\Tilde{X}^*.append(segment\_sum(X^*,J)) $ \tcp*{Base aggregate functions}
$\Tilde{X}^*.append(segment\_mean(X^*,J)) $\;
$\Tilde{X}^*.append(segment\_min(X^*,J)) $\;
$\Tilde{X}^*.append(segment\_max(X^*,J)) $\;
$\Tilde{X}^* = concatenate(\Tilde{X}^*)$ \tcp*{Combine the aggregates}
$\Tilde{X} = \Lambda_{out}(\Tilde{X}^*)$ \tcp*{Feature selection layer}
\end{algorithm}

\subsection{Data preprocessing}
\label{sec:data_process}

\begin{algorithm}
\SetAlgoLined
\caption{Database converter}
\label{algo:db2py}
\KwData{database context $DB$}
\KwResult{Data $X$ with $x \in X$ represented as $x = [x_{data},x_{ids}]$, target values $Y$, aggregation plan $P$}
$DB^* = preprocess(DB)$ \tcp*{preprocess values of tables: scale numeric data and binarize categorial data}
$P = generate\_aggregation\_plan(DB^*)$ \tcp*{Figure \ref{fig:agg_plan} shows the relation of $P$ and $DB$}
$X = []$\;
$Y = []$\;
\ForAll{$data\_instance \in DB.target\_table$}{
$Y.append(data\_instance.target)$\;
$x_{data} = [[data\_instance],[],\dots[]]$ \tcp*{The first $x_{data}$ entry is the target table}
$x_{ids} = [[],[],\dots[]]$\;
\ForAll{$nexts,current \in P$}
{
\ForAll{$next \in nexts$}
{
$next\_entries = where\_connected(x_{data}[current],DB.next)$\;
$x_{data}[next] \mathrel{+}= next\_entries$ \tcp*{Add the connected entries of table $next$}
$x_{ids}[next] \mathrel{+}= next\_entries.identifier$ \tcp*{Store the connections between the data entries}
}
}
$X.append([x_{data},x_{ids}])$\;
}
$P = invert(P)$\;
\end{algorithm}

Given a database $DB$ in star schema, a target table $DB.target\_table$ and a target attribute $DB.target\_table.target$, we transform $DB$ into the data instances $X$ and target values $Y$ and generate the aggregation plan $P$. As shown in Algorithm \ref{algo:db2py}, we first transform the values of the relational tables into an usable format by scaling and binarizing the table entries. Additionally, embeddings could also be incorporated in this step for more complex raw data like texts. Before the data set is constructed, the aggregation plan $P$ is generated by a breadth-first search through the database structure. During this search the connections between tables are memorized in tuples $p \in P$ with $p = (nexts,current)$, where $current$ is the description of the currently visited table and $nexts$ is a list of previously unvisited tables connected to the current table. After the data transformation, the order of $P$ is inverted, so that the tables farthest away from the target table get aggregated first and all data entries collapse into a single feature vector. The relation of the aggregation plan to the database structure is depicted in Figure \ref{fig:agg_plan}.

Then, for each entry in the target table, an instance in the output data set is generated. Here the target attribute is saved as $y$ and added to $Y$ and the database is traversed according to $P$. For all entries of the tables in $nexts$ that are connected to the previously selected entries of the current table, those entries are selected into the relation of the data instance and an identifier is stored to correctly map the entries for aggregation. After a complete pass through $P$, the instance $x = [x_{data},x_{ids}]$ is added to the processed data set $X$.

Our data processing approach is based on the database connection and context capabilities of the python-rdm package and therefore easy to deploy along other algorithms and functionalities provided by that package.

\begin{figure}[!htbp]
    \centering
    \resizebox{\textwidth}{!}{
    \begin{tikzpicture}[thick,]
  \matrix(users)[vtab]{users\\};
  \matrix(u2base)[right= of users,vtab]{u2base\\};
  \matrix(movies)[above= of u2base,vtab]{movies\\};
  \matrix(movies2act)[left= of movies,vtab]{movies2actors\\};
  \matrix(act)[above= of movies2act,vtab]{actors\\};
  \matrix(movies2direct)[right= of movies,vtab]{movies2directors\\};
  \matrix(direct)[above= of movies2direct,vtab]{directors\\};
    \path[->](u2base)edge(users);
    \path[->](movies)edge(u2base);
    \path[->](movies2act)edge(movies);
    \path[->](act)edge(movies2act);
    \path[->](movies2direct)edge(movies);
    \path[->](direct)edge(movies2direct);
    
    \node(transform)[below= 0.5cm of u2base]{$\Downarrow$};
    \node(P)[below= 0.5cm of transform]{
    \begin{tabular}{rl}
        $P$ =&[([actors], movie2actors), ([directors], movies2directors), ([movies2actors, movies2directors], movies),\\
         & ([movies], u2base), ([u2base], users)]
    \end{tabular}
    };

\end{tikzpicture}
}
    \caption{Relation of the aggregation plan $P$ for the MovieLens database.}
    \label{fig:agg_plan}
\end{figure}

 \subsection{Feature extraction}
 \label{sec:feature}
 
 After we have trained an N-RELAGGS model, it is able to transform a multi-relational data instance into a single predicted value or, by extracting an intermediate layer of the model, into a fixed size tensor representation. Those representations can be seen as an embedding of multi-relational data into a space of fixed dimension and therefore can be used to propositionalize the relational data and use the transformed data with arbitrary propositional learning algorithms. 
\section{Experiments}
\label{sec:experiments}

In this section we present the data sets used in our experiments as well as the performance we could measure on those. We assembled a variety of different data sets of varying size. 

Since most commonly used benchmark data sets for relational data mining are rather small, we constructed an additional data set based on the dblp citation network. The data set and the experiments conducted on it are presented in Section \ref{sec:dblp}.

\subsection{Data}

\paragraph{}
The statistics for the different databases are presented in Table \ref{tab:data}. For each database, all contained tables with their number of columns and rows are given. Additionally, for the target tables, the target attribute as well as the distribution of the different classes are specified.

\paragraph{Trains}
The Trains \cite{Trains} data set was constructed for the East-West trains challenge problem. In the challenge the direction of each train has to be predicted, based on the properties of the cars. Each train contains a variable number of cars and each has a shape and carries a load. The data set is publicly available in the Relational Dataset Repository\footnote{https://relational.fit.cvut.cz/}.

\paragraph{Mutagenesis} 
The data set \cite{Mutagenesis} comprises of 230 different chemical compounds. The task for this problem is to predict the mutagenicity of the different compounds. The data set is split into two distinct sets, one of 42 and one of 188 items, and is available in the Relational Dataset Repository.

\paragraph{Carcinogenesis} 
In this task \cite{Carcinogenesis}, the carcinogenicity of different chemical compounds has to be predicted, based on the supplied structure of the compounds. This data set is also publicly available in the Relational Dataset Repository.

\paragraph{IMDb Top} 
The IMDb Top\footnote{http://kt.ijs.si/janez\_kranjc/ilp\_datasets/imdb\_top.sql} data set is a subset of the internet movie database (IMDb) with the goal to classify the 166 selected movies either as a member of the IMDb top-250 chart or as a member of the IMDb bottom-100 chart. 

\paragraph{MovieLens} 
This \cite{MBN} is another similar database to IMDb, but for this task, the gender of the users has to be predicted. This database is publicly available in the Relational Dataset Repository.

\paragraph{Students} 
Here we want to predict whether a student will be successful in a particular study program based on the relation of the student to the study, the first semesters they are enrolled and the exams they take that semester. Since the data is sensitive, the database is not publicly available.

\begin{table}[!htbp]
    \centering
    \begin{tabular}{c|c|ccc}
         Database & tables & \#columns & \#rows & target \\
         \hline
         \hline
         \multirow{2}{*}{Trains} & cars & 10 & 63 & \\
         & trains & 2 & 20 & direction: east(10), west(10)\\
         \hline
         \multirow{6}{*}{Mutagenesis 42/188} & atoms & 5 & 1001/4893 & \\
         & bonds & 5 & 1066/5243 & \\
         & drugs & 7 & 42/188 & active: 1(13/125), 0(29/63)\\
         & rings & 2 & 259/1317 &\\
         & ring\_atom & 3 & 1785/9310 & \\
         & ring\_strucs & 3 & 279/1433 & \\
         \hline
         \multirow{6}{*}{Carcinogenesis} & atom & 5 & 8855 & \\
         & canc & 2 & 329 & class: 1(182), 0(147)\\
         & sbond\_1 & 4 & 13340 &\\
         & sbond\_2 & 4 & 940 &\\
         & sbond\_3 & 4 & 12 & \\
         & sbond\_7 & 4 & 4094 & \\
         \hline
         \multirow{7}{*}{IMDb Top} & actors & 4 & 7118 & \\
         & directors & 3 & 130 &\\
         & directors\_genres & 4 & 1123 &\\
         & movies & 4 & 166 & quality: 1(122), 0(44)\\
         & movies\_directors & 3 & 180 & \\
         & movies\_genres & 3 & 408 & \\
         & roles & 4 & 7738 & \\
         \hline
         \multirow{7}{*}{Movielens} & actors & 3 & 99129 & \\
         & directors & 3 & 2201 &\\
         & movies & 5 & 3832 &\\
         & movies2actors & 3 & 152532 &\\
         & movies2directors & 3 & 4141 & \\
         & u2base & 3 & 946828 & \\
         & users & 4 & 6039 & u\_gender: F(1708), M(4331)\\
         \hline
         \multirow{6}{*}{Student} & semester2exam & 5 & 5929 & \\
         & students & 4 & 1857 &\\
         & studies & 3 & 4 &\\
         & study2semester & 6 & 1976 &\\
         & target & 5 & 1980 & target: 1(900), 0(1080)\\
    \end{tabular}
    \caption{Statistics of the different databases used for the experiments. For each table, the number of columns and rows as well as the distribution of the target attribute for the target table are given.}
    \label{tab:data}
\end{table}

\subsection{Setup}

\paragraph{} 
The databases are loaded with the python-rdm package and split into train and test sets for two ten-fold stratified cross-validations. Those data sets are then fed into the Python wrappers of the baseline algorithms, provided by the python-rdm package, resulting in different propositional representations of those data sets. Additionally, the relational data structure is converted into numpy arrays, as described in section \ref{sec:data_process} and passed to the N-RELAGGS model as well as propositionalized by our own RELAGGS implementation. 

A feed-forward neural network of the same structure and with the same set of possible hyperparameters as the predictor part of the N-RELAGGS model is used as predictor for the different propositionalization techniques. The optimal hyperparameters, see Table \ref{tab:hyperparam}, for each predictor and the N-RELAGGS model are selected by the highest average area under the receiver operating characteristic curve (AUROC) in a stratified three-fold cross-validation over the train set. The neural networks are trained for 100 epochs, with the option of stopping early in the case of stagnating loss improvement, and we use hinge loss and Adam \cite{network:adam} to optimize the networks. For the comparison algorithms all hyperparameters are set to the default values.

In addition to the regular N-RELAGGS,  we conduct the experiments with a fixed N-RELAGGS (Fix N-RELAGGS), for whom $\lambda_{\text{feature generation}}$ and $\lambda_{\text{feature selection}}$ is fixed to $1.0$. This makes the Fix N-RELAGGS more comparable to the RELAGGS algorithm.

The experiments are executed on computing nodes with Intel Xeon CPUs, no GPUs and up to 380GB of RAM.

\begin{table}[!htbp]
    \centering
    \begin{tabular}{c|c|c}
         Algorithm & Parameter & Values  \\
         \hline
         \multirow{2}{*}{N-RELAGGS}& feature generation factor & 0.5, 0.75, 1.0 \\
         & feature selection factor & 0.5, 0.75, 1.0 \\
         \hline
         Predictor network & layer sizes & (50,), (100,), (100,50)
         
    \end{tabular}
    \caption{The different hyperparameter values for the N-RELAGGS model and the predictor networks.}
    \label{tab:hyperparam}
\end{table}

\subsection{Results}
\label{sec:results}

\begin{table}[!htbp]
    \centering
    \resizebox{\textwidth}{!}{
    \begin{tabular}{c|cccccc}
         Algorithm & Trains & Mutagenesis 188 & Mutagenesis 42 & Carcinogenesis & IMDb Top & Movielens  \\
         \hline
         Majority vote  & 0.50 (0.00) & 0.665 (0.023) & 0.695 (0.088) & 0.553 (0.012) & 0.735 (0.025) & 0.717 (0.017) \\
         \hline
Aleph  & 0.60 (0.37) & 0.744 (0.090) & 0.785 (0.192) & 0.477 (0.024) & 0.561 (0.120) & - \\
RSD  & 0.55 (0.27) & 0.846 (0.073) & \textbf{ 0.820 (0.173) } & 0.578 (0.088) & 0.699 (0.073) & - \\
Treeliker  & 0.65 (0.32) & 0.862 (0.078) & 0.785 (0.156) & 0.558 (0.066) & 0.813 (0.091) & - \\
Wordification  & 0.42 (0.33) & 0.806 (0.091) & 0.705 (0.205) & \textbf{ 0.597 (0.082) } & 0.607 (0.130) & 0.717 (0.000) \\
RELAGGS  & \textbf{ 0.70 (0.24) } & 0.840 (0.072) & 0.725 (0.211) & 0.570 (0.074) & \textbf{ 0.853 (0.080) } & 0.711 (0.066) \\
Fix N-RELAGGS  & 0.62 (0.31) & \textbf{ 0.886 (0.063) } & 0.748 (0.253) & 0.593 (0.069) & 0.833 (0.086) & 0.743 (0.032) \\
N-RELAGGS  & 0.60 (0.25) & 0.867 (0.068) & 0.808 (0.217) & 0.577 (0.063) & 0.813 (0.095) & \textbf{ 0.750 (0.035) } 
    \end{tabular}
    }
    \caption{Classification accuracy for the different algorithms and databases. For each algorithm and data pair, the average accuracy and the standard deviation over the two ten-fold cross-validations are reported. Results marked with - are omitted because the run-time exceeded 24 hours.}
    \label{tab:results_acc}
    
\end{table}

\begin{table}[!htbp]
    \centering
    \resizebox{\textwidth}{!}{
    \begin{tabular}{c|cccccc}
         Algorithm & Trains & Mutagenesis 188 & Mutagenesis 42 & Carcinogenesis & IMDb Top & Movielens  \\
         \hline
Aleph  & 0.68 (0.43) & 0.803 (0.074) & \textbf{ 0.787 (0.239) } & 0.533 (0.034) & 0.572 (0.157) & - \\
RSD  & 0.55 (0.50) & 0.946 (0.043) & 0.738 (0.349) & 0.605 (0.085) & 0.839 (0.087) & - \\
Treeliker  & 0.65 (0.48) & 0.918 (0.061) & 0.746 (0.326) & 0.589 (0.075) & 0.870 (0.092) & - \\
Wordification  & 0.55 (0.50) & 0.861 (0.076) & 0.583 (0.344) & \textbf{ 0.642 (0.087) } & 0.726 (0.128) & 0.544 (0.024) \\
RELAGGS  & \textbf{ 0.70 (0.46) } & 0.936 (0.050) & 0.629 (0.433) & 0.614 (0.068) & \textbf{ 0.893 (0.097) } & 0.667 (0.120) \\
Fix N-RELAGGS  & 0.65 (0.48) & \textbf{ 0.959 (0.052) } & 0.758 (0.317) & 0.622 (0.064) & 0.884 (0.103) & \textbf{ 0.720 (0.043) } \\
N-RELAGGS  & 0.65 (0.48) & 0.951 (0.038) & 0.779 (0.308) & 0.619 (0.083) & 0.879 (0.112) & 0.711 (0.070) 
    \end{tabular}
    }
    \caption{AUROC scores for the different algorithms and databases. The reported values are the average and the standard deviation over the two ten-fold cross-validations. Results marked with - are omitted because the run-time exceeded 24 hours.}
    \label{tab:results_roc}
\end{table}

\begin{table}[!htbp]
    \centering
    \begin{tabular}{l|c||l|c||l|c}
    \multicolumn{2}{c||}{Accuracy} & \multicolumn{2}{c||}{AUROC} & \multicolumn{2}{c}{Combined} \\
    \hline
    Algorithm & Rank & Algorithm & Rank & Algorithm & Rank \\
\hline
        Fix N-RELAGGS & 2.33 &Fix N-RELAGGS & 2.21 &Fix N-RELAGGS & 2.27 \\
N-RELAGGS & 2.88 &N-RELAGGS & 2.62 &N-RELAGGS & 2.75 \\
RELAGGS & 3.67 &RELAGGS & 3.33 &RELAGGS & 3.50 \\
Treeliker & 4.08 &Treeliker & 4.38 &Treeliker & 4.23 \\
RSD & 4.83 &RSD & 4.92 &RSD & 4.88 \\
Wordification & 5.38 &Wordification & 5.29 &Wordification & 5.33 \\
Majority vote & 6.12 &Aleph & 5.79 &Aleph & 6.25 \\
Aleph & 6.71 &Majority vote & 7.46 &Majority vote & 6.79 

    \end{tabular}
    \caption{Average ranks of the algorithms based on the performance in accuracy, AUROC and the combination of both.}
    \label{tab:mean_ranks}
\end{table}

\begin{table}[!htbp]
    \centering
    \resizebox{\textwidth}{!}{
    \begin{tabular}{rcl|rcl|rcl}
    \multicolumn{3}{c}{Accuracy} & \multicolumn{3}{c}{AUROC} & \multicolumn{3}{c}{Combined}\\
    \hline
        RELAGGS&$>$&Aleph&RELAGGS&$>$&Majority vote&RELAGGS&$>$&Aleph\\
N-RELAGGS&$>$&Aleph&N-RELAGGS&$>$&Aleph&RELAGGS&$>$&Majority vote\\
N-RELAGGS&$>$&Majority vote&N-RELAGGS&$>$&Majority vote&N-RELAGGS&$>$&Aleph\\
Fix N-RELAGGS&$>$&Aleph&Fix N-RELAGGS&$>$&Aleph&N-RELAGGS&$>$&RSD\\
Fix N-RELAGGS&$>$&Wordification&Fix N-RELAGGS&$>$&Wordification&N-RELAGGS&$>$&Wordification\\
Fix N-RELAGGS&$>$&Majority vote&Fix N-RELAGGS&$>$&Majority vote&N-RELAGGS&$>$&Majority vote\\
& & & Treeliker&$>$&Majority vote&Fix N-RELAGGS&$>$&Aleph\\
& & & & & & Fix N-RELAGGS&$>$&RSD\\
& & & & & & Fix N-RELAGGS&$>$&Wordification\\
& & & & & & Fix N-RELAGGS&$>$&Majority vote\\
& & & & & & Treeliker&$>$&Majority vote
    \end{tabular}
    }
    \caption{Statistical significant differences of the algorithms based on their ranks for the performance measures. We use the Bonferroni-Dunn test with $\alpha = 0.05$ to prove the significance.}
    \label{tab:algo_better}
\end{table}

\begin{figure}[!htbp]
    \centering
    \begin{subfigure}[b]{\textwidth}
    \centering
    \resizebox{0.8\textwidth}{!}{
    \includegraphics{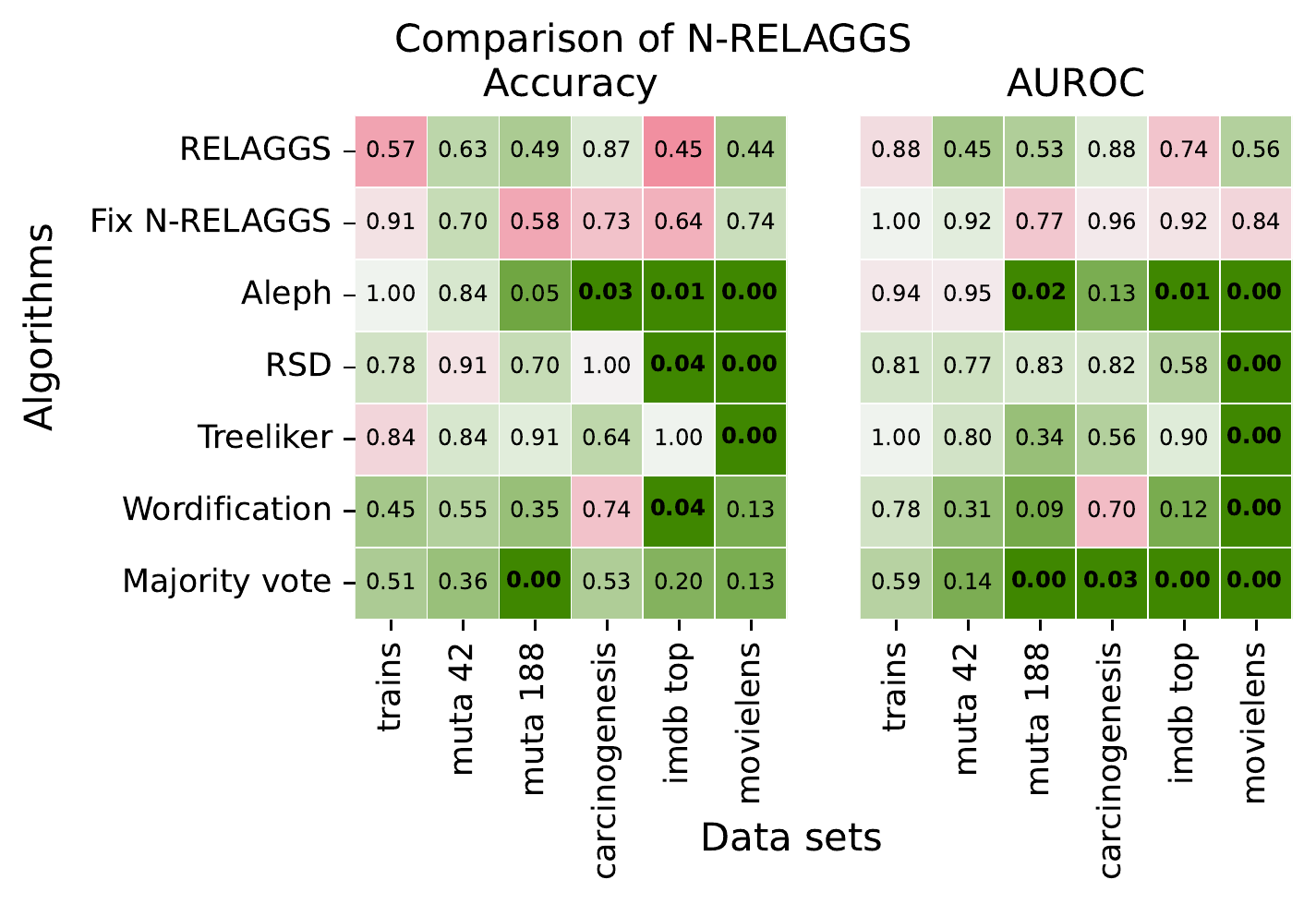}
}
    \caption{N-RELAGGS}
    \label{fig:nrelaggs_diff}
    \end{subfigure}
    \hfill
     \begin{subfigure}[b]{\textwidth}
    \centering
    \resizebox{0.8\textwidth}{!}{
    \includegraphics{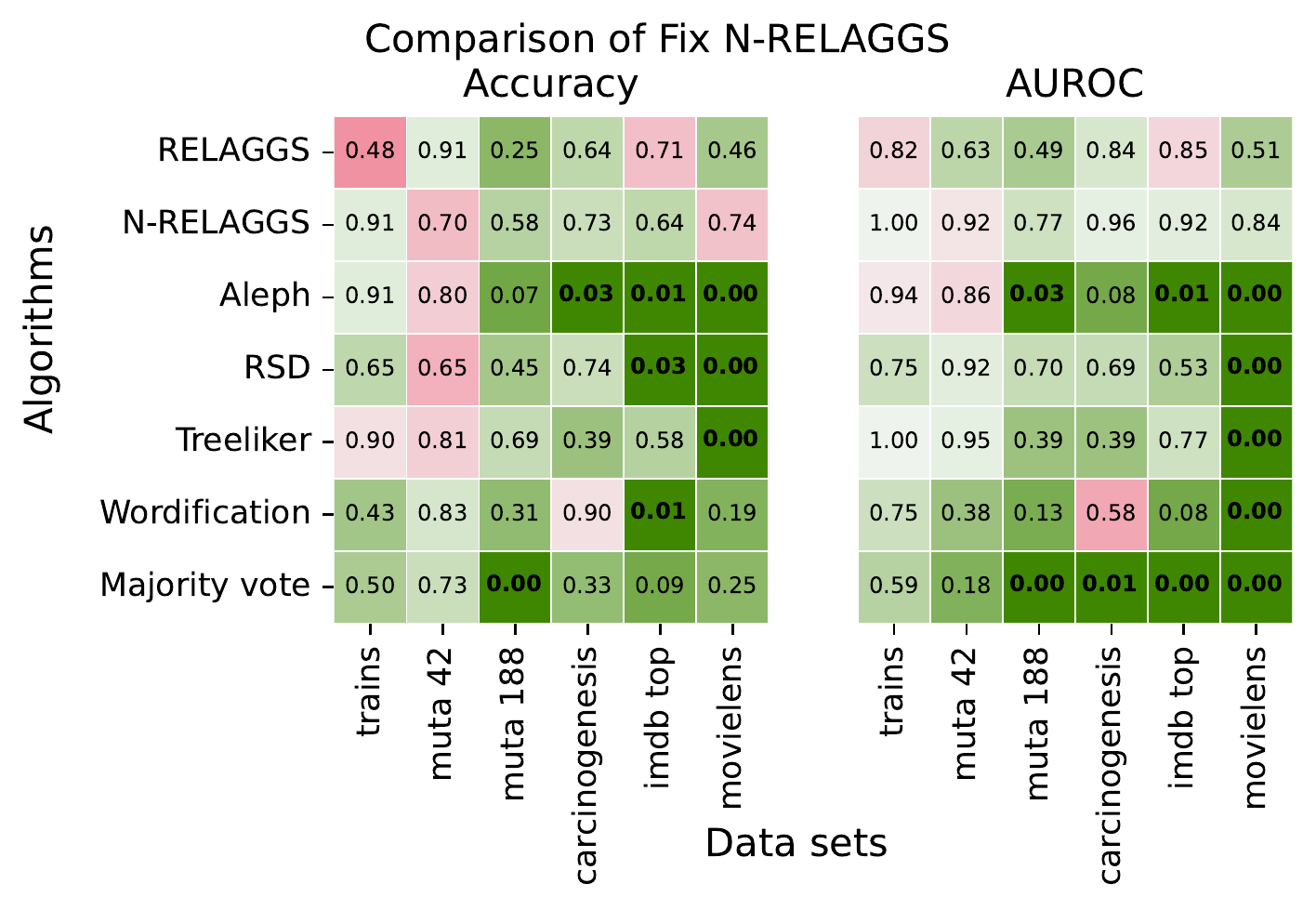}
}
    \caption{Fix N-RELAGGS}
    \label{fig:fixnrelaggs_diff}
    \end{subfigure}
    \caption{Comparison between the N-RELAGGS algorithm, respectively the Fix N-RELAGGS algorithm, with all other tested algorithms. Green fields show superiority, red fields show inferiority and the saturation of the fields represents the significance of the difference. Additionally each field yields the p-value determined by a corrected repeated 10-fold cv test with two repetitions and the bar for significance set to $\alpha = 0.05$.}
    \label{fig:all_diff}
\end{figure}

The experimental results of the selected algorithms on the different data sets are presented in this section. Table \ref{tab:results_acc} yields the measured accuracies and Table \ref{tab:results_roc} the measured AUROC scores for the different algorithm and data pairs. The average ranks of the algorithms based on their performances is presented in Table \ref{tab:mean_ranks}.

Using the ranks of the algorithms we can test, if there are significant differences in the performances of the algorithms. Utilizing an adapted Friedman test \cite{ranking:signific} we can prove, that there are significant ($p < 0.001$) differences in the distribution of ranks, i.e. the quality of the algorithms. To determine significant differences between the algorithms we use the Bonferri-Dunn test \cite{ranking:signific} with a significance threshold of $\alpha = 0.05$ and present the results in Table \ref{tab:algo_better}.

Figure \ref{fig:all_diff} shows for which data sets the N-RELAGGS or Fix N-RELAGGS algorithm performs better or worse than the other algorithms, as well as the significance of the difference indicated by the p-value determined by a corrected repeated 10-fold cv test \cite{corrTtest} with two repetitions.

\subsection{DBLP data set}
\label{sec:dblp}
This database is constructed from the DBLP-Citation-network V12\footnote{https://originalstatic.aminer.cn/misc/dblp.v12.7z} data set \cite{dblp_data}, in order to build a large benchmark data set. The instances are the papers published between 2010 and 2012 with a threshold for the number of citations used as target. The positive class 1 consists of all papers with more than 10 citations and the negative class 0 of all papers with 10 or fewer citations. In order to prevent leakage, the database table all\_references is solely composed of papers published prior to 2010. Table \ref{tab:data_dblp} presents the overall statistics of the DBLP database\footnote{https://github.com/kramerlab/n-relaggs/Data/DBLP}.

Because of the size of the database, most algorithms are not able to propositionalize the database in five days, which is the longest possible run time on the compute nodes. Just our transformation for the RELAGGS and N-RELAGGS algorithms is able to complete the data set. This is primarily due to the potential for parallelization and integrated checkpointing, which enables the algorithm to transform multiple small batches of the database independently on different nodes as well as continuing the computation after termination with a minimal overhead.

Table \ref{tab:dblp-perf} yields the performance of the experiments on the DBLP data set. The presented results are the average values of the performance scores of two runs with a random sampled test set of 224199 instances on the data set. For now, just one set of hyperparameters is used. The classifier has one hidden layer with 100 neurons and $\lambda_{\text{feature genertaion}}$ and $\lambda_{\text{feature selection}}$ are set to $1.0$. 

\begin{table}[!htbp]
    \centering
    \begin{tabular}{c|ccc}
         tables & \#columns & \#rows & target \\
         \hline
         \hline
         all\_references & 5 & 1639164 & \\
         author2paper & 3 & 2787326 &\\
         org2paper & 2 & 888618 &\\
         paper2author & 3 & 1386766 &\\
         paper2org & 2 & 473104 &\\
         paper2paper & 2 & 5984894 &\\
         papers & 5 & 724199 & target:  0(504754), 1(219445) \\
    \end{tabular}
    \caption{Statistics of the DBLP database used for the experiments. For each table, the number of columns and rows as well as the distribution of the target attribute for the target table are given.}
    \label{tab:data_dblp}
\end{table}

\begin{table}[!htbp]
    \centering
    \begin{tabular}{c|cc}
        Algorithm & Accuracy & AUROC \\
        \hline
        Majority vote  & 0.697 (0.001) & 0.500 (0.000) \\
        \hline
RELAGGS  & 0.732 (0.003) & 0.630 (0.013) \\
N-RELAGGS  & \textbf{ 0.739 (0.002) } & \textbf{ 0.668 (0.003) }
    \end{tabular}
    \caption{Scores for experiments conducted on the DBLP database. The scores are the average of two experiments runs with 500000 ($\sim 69\%$) randomly selected instances as training set and 224199 ($\sim 31\%$) instances as test set.}
    \label{tab:dblp-perf}
\end{table}

\subsection{Feature extraction}
\label{sec:feat_results}
As mentioned in Section \ref{sec:feature}, we can extract intermediate layers of the N-RELAGGS model and use them to transform relational data instances into single data vectors. This propositionalized representation can then be used as input for other algorithms, for instance, we can compare the performance of the different propositionalizations when classified using a random forest classifier, as shown in Table \ref{tab:decision_tree_roc}. This experiment is run with the scikit-learn \cite{scikit-learn} implementation of the random forest classifier and all parameters set to the default values.

\begin{table}[!htbp]
    \centering
    \begin{tabular}{c|ccccccc}
         Algorithm & Trains & Mutagenesis 188 & Mutagenesis 42 & Carcinogenesis  \\
         \hline
         Aleph & 0.90 (0.12) & 0.665 (0.009) & 0.950 (0.100) & 0.553 (0.007)  \\ 
         RSD & 0.75 (0.16) & 0.767 (0.064) & 0.811 (0.088) & 0.6088 (0.030) \\ 
         Treeliker & 0.75 (0.27) & 0.660 (0.009) & 0.833 (0.058) & 0.596 (0.019) \\ 
         Wordification & 0.70 (0.24) & 0.676 (0.010) & 0.833 (0.058) & 0.608 (0.056)  \\ 
         RELAGGS & 0.75 (0.00) & 0.872 (0.030)& 0.836 (0.116) & 0.620 (0.054)  \\ 
         N-RELLAGS & 0.65 (0.20) & 0.910 (0.046) & 0.856 (0.095) & 0.611 (0.033)  \\ 
    \end{tabular}
    \caption{Accuracy for the different algorithms and databases when classified by a random forest. The reported values are the average and the standard deviation over the five-fold cross-validation.}
    \label{tab:decision_tree_roc}
\end{table}

\section{Discussion}
\label{sec:discussion}

The results presented in section \ref{sec:results} show the superior predictive performance of our proposed N-RELAGGS algorithm, compared to other state-of-the-art algorithms. The average ranks presented in Table \ref{tab:mean_ranks} already show, that N-RELAGGS is among the best performing algorithms, but with Table \ref{tab:algo_better} we prove that N-RELAGGS performs significantly better than most compared algorithms and that no other algorithm performs significantly better than N-RELAGGS.

Considering the comparisons presented in Figure \ref{fig:all_diff}, we see that the performance regarding the AUROC of N-RELAGGS is significantly better than all other algorithms, except for the regular RELAGGS. Combining the performance differences in Table \ref{tab:winLoss} shows that the number of positive comparisons is at least double the number of negative comparisons for all data sets except trains.

\begin{table}[!htbp]
    \centering
    \begin{tabular}{c|ccc|ccc}
        & \multicolumn{3}{c|}{N-RELAGGS} & \multicolumn{3}{c}{Fix N-RELAGGS} \\
        Data set & better & worse & equal & better & worse & equal \\
        \hline
        trains & 6 & 4 & 2 & 7 & 4 & 1 \\
        muta 42 & 10 & 2 & 0 & 8 & 4 & 0 \\
        muta 188 & 12(3) & 0 & 0 & 12(3) & 0 & 0 \\
        carcinogenesis & 9(2) & 2 & 1 & 10(2) & 2 & 0 \\
        imdb top & 9(5) & 2 & 1 & 10(5) & 2 & 0 \\
        movielens & 12(8) & 0 & 0 & 12(8) & 0 & 0
    \end{tabular}
    \caption{Differences in predictive performance of N-RELAGGS/ Fix N-RELAGGS compared to the other algorithms on the data sets. Each tuple represents better performance, worse performance and equal performance with the numbers in brackets representing the number of significant differences.}
    \label{tab:winLoss}
\end{table}

In Section \ref{sec:dblp} we introduce the very large DBLP database and try to run all algorithms on it. So far only our implementation was able to complete a run on the database, due to its high parallelizability. The results presented in Table \ref{tab:dblp-perf} show the superior performance of N-RELAGGS compared to RELAGGS.

Comparing RELAGGS to N-RELAGGS, especially Fix N-RELAGGS which resembles the structure of RELAGGS more closely, gives no statistically significant difference. Although RELAGGS outperforms N-RELAGGS just on two of six data sets, it ranks on average below N-RELAGGS, which suggests that N-RELAGGS peforms better than RELAGGS.

We show that N-RELAGGS is a regular propositionalization algorithm with the experiment in Section \ref{sec:feat_results}.
We train the network independent of the actual predictor and select an intermediate layer as propositional representation of the database. This way we can transform the whole relational database into a propositional representation, i.e. propositionalize the database.
Table \ref{tab:decision_tree_roc} suggests that the predictive performance of the N-RELAGGS propositionalization is similar to the performance of other algorithms. In terms of the sum of ranks, RELAGGS is slightly superior to N-RELAGGS, however, in the direct comparison, each ``win'' on two data sets each. No attempts have been made yet to optimize the hyperparameters of N-RELAGSS with respect to the subsequent random forest classifier.

\section{Conclusion and Future Work}
\label{sec:conclusion}

In this paper we introduce the idea of composite aggregate functions as a way to build a trainable aggregate function from a simple unparameterized aggregate function and two parameterized mappings, and we present a neural network based implementation of them. By combining multiple of those aggregation layers in a network architecture similar to the aggregation scheme of the RELAGGS algorithm, we construct the N-RELAGGS network.

We compare the predictive performance of our proposed N-RELAGGS algorithm with multiple baseline methods, including the structurally similar RELAGGS algorithm, on seven relational data sets and show the advantage of trainable aggregate functions over static aggregate functions.
Additionally we present the DBLP database as a large benchmark data set for propositionalization and relational learning.

A possibility of future improvements is the incorporation of other base aggregate functions into the network architecture. One such possible option is the addition of recurrent neural networks as aggregate functions \cite{sogood}.

\section*{Acknowledgments}
This research has been partially funded by the Federal Ministry of Education and Research of Germany in the framework of Lehren, Organisieren, Beraten: Gelingensbedingungen von Bologna (LOB) (project number 01PL17055).

Parts of this research were conducted using the supercomputer Mogon and/or advisory services offered by Johannes Gutenberg University Mainz (hpc.uni-mainz.de), which is a member of the AHRP (Alliance for High Performance Computing in Rhineland Palatinate,  www.ahrp.info) and the Gauss Alliance e.V.

The authors gratefully acknowledge the computing time granted on the supercomputer Mogon at Johannes Gutenberg University Mainz (hpc.uni-mainz.de).

\bibliographystyle{unsrt}  
\bibliography{my_bib}

\end{document}